\documentclass[10pt,twocolumn,letterpaper]{article}

\usepackage[pagenumbers]{cvpr} 
\usepackage{url}
\usepackage{mdframed}
\usepackage{listings}
\usepackage{xcolor}
\usepackage{booktabs}
\usepackage{multirow}
\usepackage{array}
\usepackage{colortbl}
\usepackage{graphicx}
\usepackage{pifont}
\usepackage[accsupp]{axessibility}

\definecolor{cvprblue}{rgb}{0.21,0.49,0.74}
\usepackage[pagebackref,breaklinks,colorlinks,allcolors=cvprblue]{hyperref}

\def\paperID{17010} 
\def\confName{CVPR}
\def\confYear{2026}
\def\approach{AnyDoc}
\def\dataset{DocHTML}

\title{\approach{}: Enhancing Document Generation via Large-Scale HTML/CSS Data Synthesis and Height-Aware Reinforcement Optimization}

\author{Jiawei Lin\textsuperscript{*}\\
Xi'an Jiaotong University\\
{\tt\small kylelin@stu.xjtu.edu.cn}
\and
Wanrong Zhu\\
Adobe Research\\
{\tt\small wzhu@adobe.com}
\and
Vlad I Morariu\\
Adobe Research\\
{\tt\small morariu@adobe.com}
\and
Christopher Tensmeyer\\
Adobe Research\\
{\tt\small tensmeye@adobe.com}
}

\begin{document}
\twocolumn[{
    \renewcommand\twocolumn[1][]{#1}
    \maketitle
    \begin{center}
        \centering
        \includegraphics[width=\textwidth]{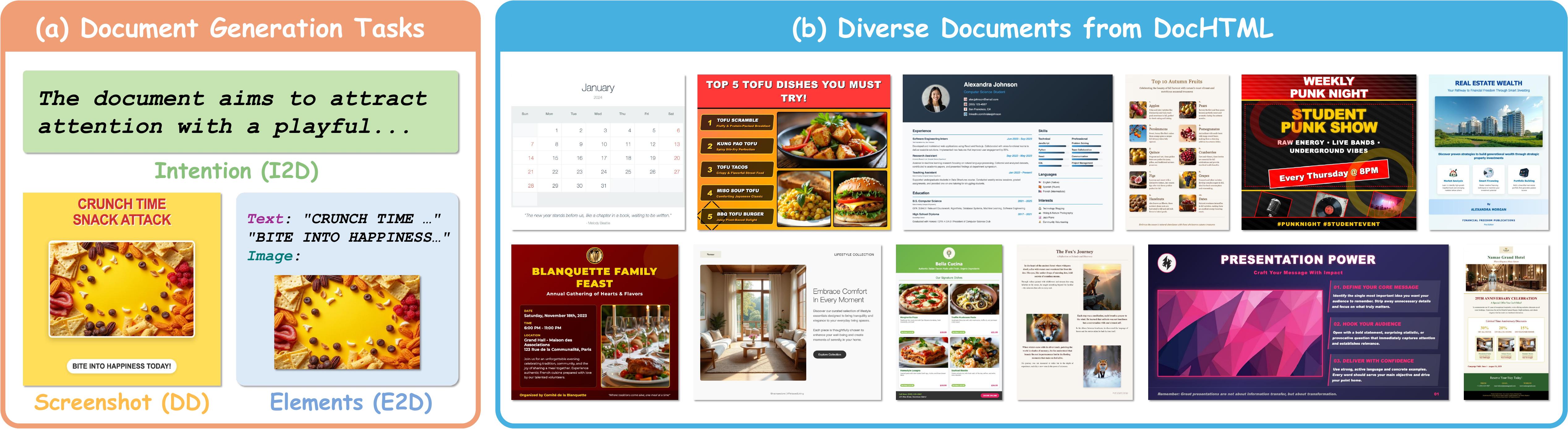}
        \captionof{figure}{(a) \approach{} excels at generating layered documents from diverse input modalities, including natural language design intentions, screenshots to be derendered, or collections of text and image elements. (b) Our curated \dataset{} covers a broad range of document categories and styles.}
        \label{fig:teaser}
    \end{center}
}]
\maketitle
\renewcommand{\thefootnote}{\fnsymbol{footnote}}                                   
\footnotetext[1]{Work done during an internship at Adobe Research.} \renewcommand{\thefootnote}{\arabic{footnote}}
\begin{abstract}
Document generation has gained growing attention in the field of AI-driven content creation.
In this work, we push its boundaries by introducing \approach{}, a framework capable of handling multiple generation tasks across a wide spectrum of document categories, all represented in a unified HTML/CSS format.
To overcome the limited coverage and scale of existing human-crafted document datasets, \approach{} first establishes a scalable data synthesis pipeline to automatically generate documents in HTML/CSS form.
This pipeline yields \dataset{}, a large-scale dataset containing 265,206 document samples, while spanning 111 categories and 32 distinct styles.
Additionally, all documents are equipped with comprehensive metadata, including design intentions, HTML/CSS source code, visual assets, and rendered screenshots.
Building on the curated dataset, \approach{} fine-tunes multi-modal large language models (MLLMs) to achieve three practical document generation tasks: intention-to-document, document derendering, and element-to-document.
To address the content overflow issue observed during fine-tuning, \approach{} further incorporates a height-aware reinforcement learning (HARL) post-training procedure.
By defining a reward function based on the difference between predicted and target document heights, overflow is penalized and gradually mitigated during HARL, thereby enhancing overall performance.
Qualitative and quantitative experiments demonstrate that \approach{} outperforms both general-purpose MLLMs and task-specific baselines across all three tasks.
\end{abstract}    
\section{Introduction}
\label{sec:intro}

We live in a world of documents.
On a daily basis, we view, edit, and share documents from various categories, such as resumes, presentations, reports, menus, letters, and more.
Compared to plain text, they offer a more structured, efficient, and aesthetically pleasing information flow through well-designed layouts, typography, and color schemes.

The creation of high-quality documents, however, remains a labor-intensive task.
It requires a synergistic balance of multiple design principles, including structural clarity, layout aesthetics, visual harmony, and stylistic coherence.
Recently, with the advancement of generative models, there has been an increasing trend toward automated document generation~\cite{chen2025posta, ge2025autopresent, zheng2025pptagent, lin2025elements, graphist2023hlg}.
Taking a design intention or a set of elements as input (Figure~\ref{fig:teaser}a), current approaches can produce compliant documents, effectively streamlining the document creation process.

\begin{figure}
    \centering
    \includegraphics[width=\linewidth]{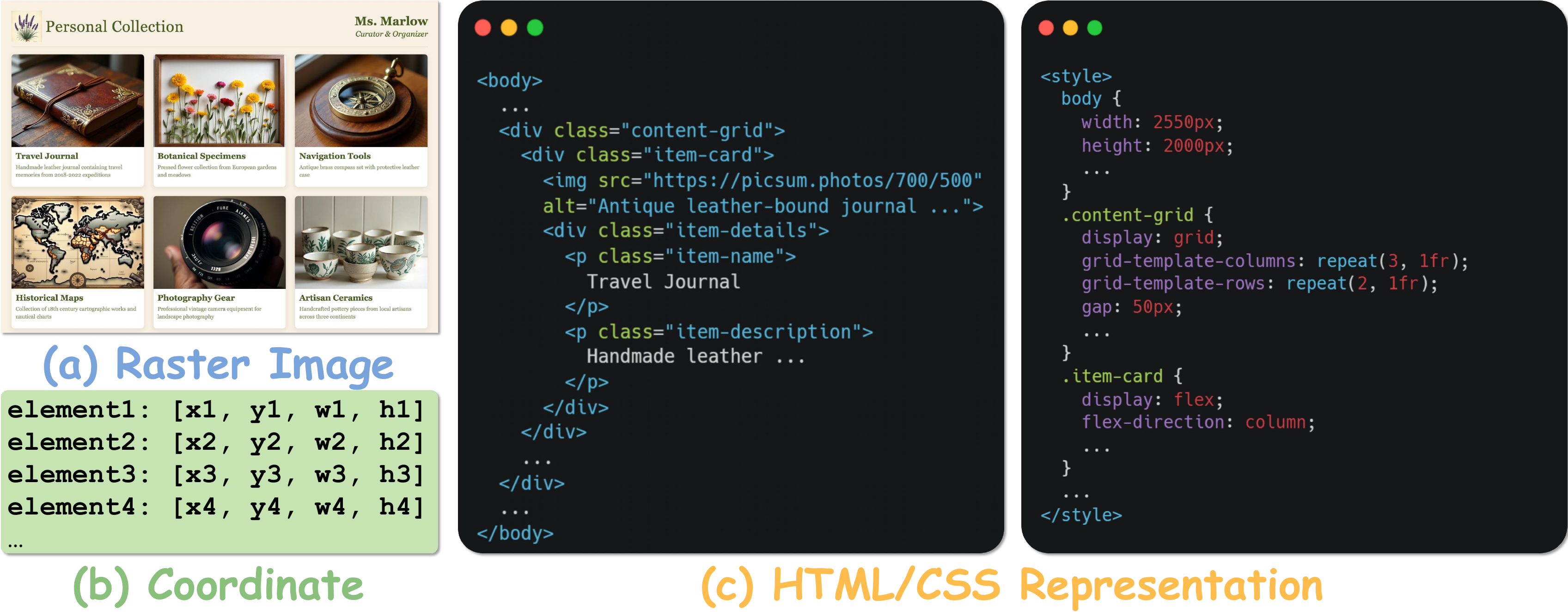}
    \caption{Existing approaches generate documents as either (a) raster images or (b) flat element coordinate sequences. (c) In \approach{}, we generate documents with a hierarchical and multi-layered HTML/CSS representation.}
    \vspace{-2mm}
    \label{fig:different_data_representation}
\end{figure}

Despite recent progress, we highlight several critical limitations of existing document generation methods, summarized as follows.
(1) \textit{Limited application scope.}
Most methods are specifically designed for a single document category, such as advertisements~\cite{lin2023autoposter, li2023planning}, presentations~\cite{zheng2025pptagent, ge2025autopresent}, or infographics~\cite{peng2025bizgen}, and have difficulty in handling unseen categories.
(2) \textit{Suboptimal document representation.}
As shown in Figure~\ref{fig:different_data_representation}, there are two widely used representations in previous work: raster image~\cite{pu2025art,peng2025bizgen,gao2025postermaker,wang2025designdiffusion} and flat coordinate sequence~\cite{inoue2023towards,graphist2023hlg,lin2025elements}, each of which suffers from notable shortcomings.
The image representation largely sacrifices document editability.
The flat sequence, though editable, struggles with structurally complex documents, as they require extensive coordinate calculations to ensure proper containment, spacing, and grouping relationships between elements.
(3) \textit{Data scarcity.}
Manually crafting documents is expensive and difficult to scale, resulting in limited size and coverage of existing datasets~\cite{yamaguchi2021canvasvae, chen2025posta, hsu2023posterlayout}.
Insufficient high-quality data prevents document generation models from reaching their full potential.

To address the above issues, our key insight is to introduce HTML/CSS (Figure~\ref{fig:different_data_representation}c), which are commonly adopted in web development, into document generation for their following advantages.
First, HTML inherently provides a well-defined hierarchical structure, making it more suitable for representing structured documents than the flat coordinate counterpart.
For example, the parent–child hierarchies in HTML naturally reflect the containment relationships among elements.
Second, CSS offers powerful layout mechanisms that enable sophisticated document designs through declarative styling.
Its predefined layout modules, such as flexbox and grid, are able to model grouping, alignment, and spacing relationships without precise numerical calculations, allowing the model to focus on high-level design intent rather than low-level coordinate management.
Third, the widespread adoption of HTML/CSS renders it possible to synthesize large-scale, multi-category documents via off-the-shelf code generation models~\cite{deepseek-coder, hui2024qwen2}, thereby addressing data scarcity as well as broadening the application scope.
Finally, HTML/CSS maintain full editability.
All elements can be freely replaced, restyled, or repositioned by editing the source code.

\begin{figure*}
    \centering
    \includegraphics[width=\linewidth]{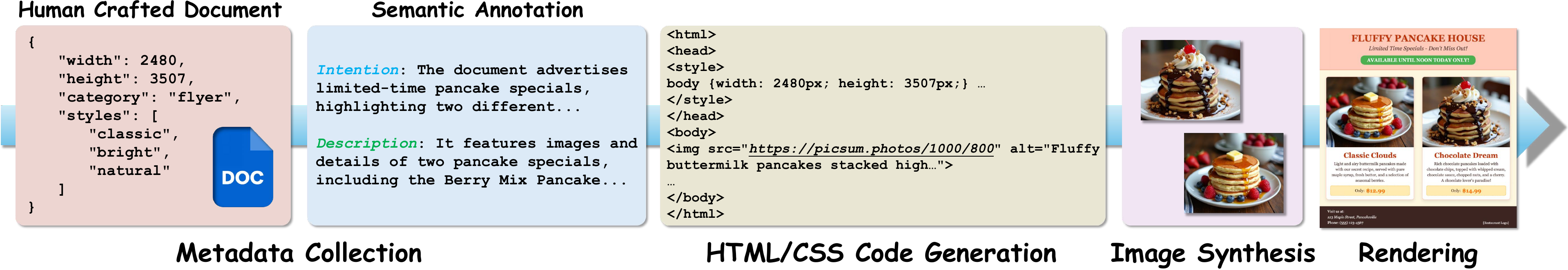}
    \caption{Overview of the HTML/CSS document synthesis pipeline. Starting from a human-crafted document and its accompanying metadata, we employ MLLMs to generate semantic annotations, i.e., the document’s intention and description. Based on the conditions, a code generation model and an image generation model are sequentially used to synthesize corresponding HTML/CSS code and image assets, which are finally rendered into the complete document screenshot.}
    \vspace{-3mm}
    \label{fig:dataset_pipeline}
\end{figure*}

Based on these insights, we present \textbf{\approach{}}, a versatile framework that leverages the unified HTML/CSS representation for automatic document generation across diverse categories.
As shown in Figure~\ref{fig:teaser}a, it considers three practical generation tasks: intention-to-document (I2D), document derendering (DD), and element-to-document (E2D).
Our approach starts by synthesizing a large-scale, wide-ranging, and high-quality dataset.
To achieve this, \approach{} establishes an automated data synthesis pipeline (Figure~\ref{fig:dataset_pipeline}) comprising five key stages: metadata collection, HTML/CSS code generation, image asset synthesis, rendering, and data cleaning.
By utilizing the state-of-the-art image and code generation models in the pipeline, we obtain a total of 265,206 HTML/CSS documents, covering 111 document categories and 32 styles (Figure~\ref{fig:teaser}b).
We refer to this dataset as \textbf{\dataset{}}.
\approach{} then fine-tunes multi-modal large language models (MLLMs) on \dataset{} to achieve strong document generation capabilities.

After fine-tuning, a post-training stage is introduced to address the overflow problem, where elements in some generated documents exceed the specified heights. 
Specifically, we propose Height-Aware Reinforcement Learning (HARL), built upon Group Relative Policy Optimization (GRPO)~\cite{shao2024deepseekmath}, to penalize over-height outputs. 
HARL utilizes Playwright to render generated HTML/CSS code into document images to obtain their actual heights. 
Rewards are then calculated based on the ratio of actual heights to specified heights.
The more overflow, the smaller the reward.
Through HARL, the model learns to comply with height constraints, further enhancing overall performance.

We compare \approach{} against both general-purpose MLLMs and specialized models.
Both quantitative and qualitative results show its superiority over baseline methods across all three tasks.
Moreover, ablation studies reveal the clear advantages of the used HTML/CSS representation, the synthesized dataset \dataset{}, and HARL.
These findings establish \approach{} as an effective solution for versatile, high-quality, and input-compliant document generation.

\section{Related Work}
\label{sec:related_work}

\noindent \textbf{Document Generation.}
With the great advancements of text-to-image diffusion models in visual fidelity and text rendering accuracy~\cite{flux2024, esser2024scaling, podell2023sdxl}, one line of existing work fine-tunes these generators for high-quality document image generation~\cite{wang2025designdiffusion, jia2023cole, inoue2024opencole, hu2025dreamposter}.
Some of them also introduce additional attention mechanisms for precise element positioning (e.g., product images, text)~\cite{gao2025postermaker, zhang2025creatidesign, peng2025bizgen}.
While these methods can produce visually compelling rendering results, the raster outputs lack editability, significantly restricting their practical applicability.
Recent approaches have partially addressed this limitation by decomposing documents into multi-layer transparent images~\cite{pu2025art, chen2025prismlayers}. 
However, they remain insufficient for comprehensive control, particularly over text content, color schemes, font selection, and other fine-grained styling attributes.

Another line of work focuses on editable document creation.
This direction spans multiple document categories, including academic posters~\cite{zhang2025postergen, pang2025paper2poster}, slides~\cite{zheng2025pptagent, ge2025autopresent}, and graphic designs~\cite{graphist2023hlg, lin2025elements}, where documents are typically represented as flat coordinate sequences (e.g., JSON).
Although these methods enable post-generation editing, they suffer from quality issues when encountering complex documents, since the flat representations introduce extensive coordinate calculations to produce proper layout structures.
In this paper, we propose \approach{}, which leverages the HTML/CSS representation to address the limitations of both lines of work.
Notably, although some previous approaches also use HTML, they merely adopt the format while still performing coordinate prediction~\cite{tang2023layoutnuwa, lin2023layoutprompter, shao2024webrpg, seol2024posterllama}, whereas we demonstrate the full potential of hierarchical HTML/CSS representation for document generation.

\vspace{2mm}
\noindent \textbf{Document Dataset Curation.}
In addition to advanced algorithms, high-quality datasets are also critical for document generation~\cite{yamaguchi2021canvasvae, chen2025posta, chen2025postercraft, chen2025prismlayers, zhang2025creatidesign}.
Among the public datasets, some rely on heavy manual processes~\cite{yamaguchi2021canvasvae, chen2025posta}, some do not provide editable document formats~\cite{zhang2025creatidesign, chen2025postercraft}, and all suffer from limited size and coverage.
For example, the widely used Crello~\cite{yamaguchi2021canvasvae} contains merely 20K graphic designs across a few categories.
To solve this, we propose an automated document synthesis pipeline and contribute a dataset \dataset{} based on it.
\dataset{} surpasses existing datasets in terms of coverage, scale, and editability.

\vspace{2mm}
\noindent \textbf{Reinforcement Learning for Large Language Models.}
Reinforcement learning techniques, such as RLHF~\cite{ouyang2022training}, DPO~\cite{rafailov2023direct}, and GRPO~\cite{shao2024deepseekmath}, have proven effective in aligning large language models with desired behaviors. 
Among these, GRPO has gained considerable attention due to its good performance and computational efficiency.
In \approach{}, we build HARL based on GRPO.
A task-specific reward is introduced in HARL to quantify content overflow. 
This enables better height control over the output documents after RL optimization.
\section{Methodology}
\label{sec:method}

\subsection{Dataset: \dataset{} Construction}
\label{sec:dataset}

As illustrated in Figure~\ref{fig:dataset_pipeline}, the data synthesis pipeline comprises several key steps:
(1) metadata collection, ensuring diverse coverage across document categories, styles, and intentions;
(2) HTML/CSS code generation, producing structured document representations;
(3) image asset synthesis, populating visual content to the image placeholder; and
(4) rendering, obtaining document screenshots.
Additionally, rigorous data cleaning rules are employed to maintain the high quality of the dataset.

\vspace{2mm}
\noindent \textbf{Metadata Collection.}
We use a large-scale, professionally crafted document repository as the metadata source.
Please refer to the supplementary materials for detailed statistics on its document categories, styles, and canvas sizes, which exhibit broad diversity.
Since the repository does not include textual metadata, we employ InternVL3~\cite{zhu2025internvl3}, an advanced multi-modal large language model, to create semantic annotations for each document.
Specifically, we prompt the model to extract two kinds of information: (1) high-level intentions that describe the document’s design purpose, and (2) detailed descriptions that objectively summarize its visual content.
The annotation prompts are also provided in the supplementary materials.

\vspace{2mm}
\noindent \textbf{HTML/CSS Code Generation.}
Using the collected metadata as input conditions, we leverage Qwen3-Coder-480B~\cite{yang2025qwen3} to generate corresponding HTML/CSS documents.
To fully exploit the expressive power of the HTML/CSS representation and ensure compatibility with subsequent steps, we introduce several key principles into the prompt.
First, we encourage the use of modern CSS layout modules (e.g., flexbox and grid) for element positioning and spacing control.
Specifically, we utilize CSS properties such as row-gap, column-gap, justify-content, and align-items for precise alignment and spacing, while flex-basis, flex-grow, and flex-shrink are applied to manage element sizing.
Second, for the \texttt{<img>} tag, we define a standardized output format (see Figure~\ref{fig:dataset_pipeline}).
The \texttt{src} field adopts a unified placeholder URL (\textit{https://picsum.photos/W/H}) to include the image width and height.
While the \texttt{alt} field is required to provide a detailed textual description of the desired image content.
Through this mechanism, it effectively enforces size coordination, semantic consistency, and contextual harmony between an image element and its surroundings.
Third, the model is instructed to set the HTML body dimensions according to the specified document width and height.
Detailed prompts are included in the supplementary materials.

\vspace{2mm}
\noindent \textbf{Image Asset Synthesis \& Rendering.}
We input the widths, heights, and text descriptions predicted in the previous step into FLUX.1-dev~\cite{flux2024} to synthesize high-quality visual assets.
After image generation, the placeholder URLs in the source code are replaced with actual image file paths.
Next, we use Playwright, a browser automation framework, to render the HTML/CSS code with synthesized images into document screenshots.
Notably, unlike raster document datasets, our pipeline provides documents in both code and visual forms, thus ensuring full editability.

\vspace{2mm}
\noindent \textbf{Data Cleaning.}
A document is excluded from the dataset if it meets any of the following conditions:
(1) the HTML body dimensions do not match the specified width and height;
(2) image tags lack the required \texttt{src} placeholder or \texttt{alt} content;
(3) elements have zero height values; or
(4) content overflows beyond the document boundaries.
After data cleaning, we obtain \dataset{}, a high-quality document dataset consisting of 265,206 samples, 111 categories, and 32 styles.
All samples have rich components (metadata, HTML/CSS source code, synthesized images, and rendered screenshots), thus establishing a solid foundation for document generation.

\begin{figure}
    \centering
    \includegraphics[width=0.95\linewidth]{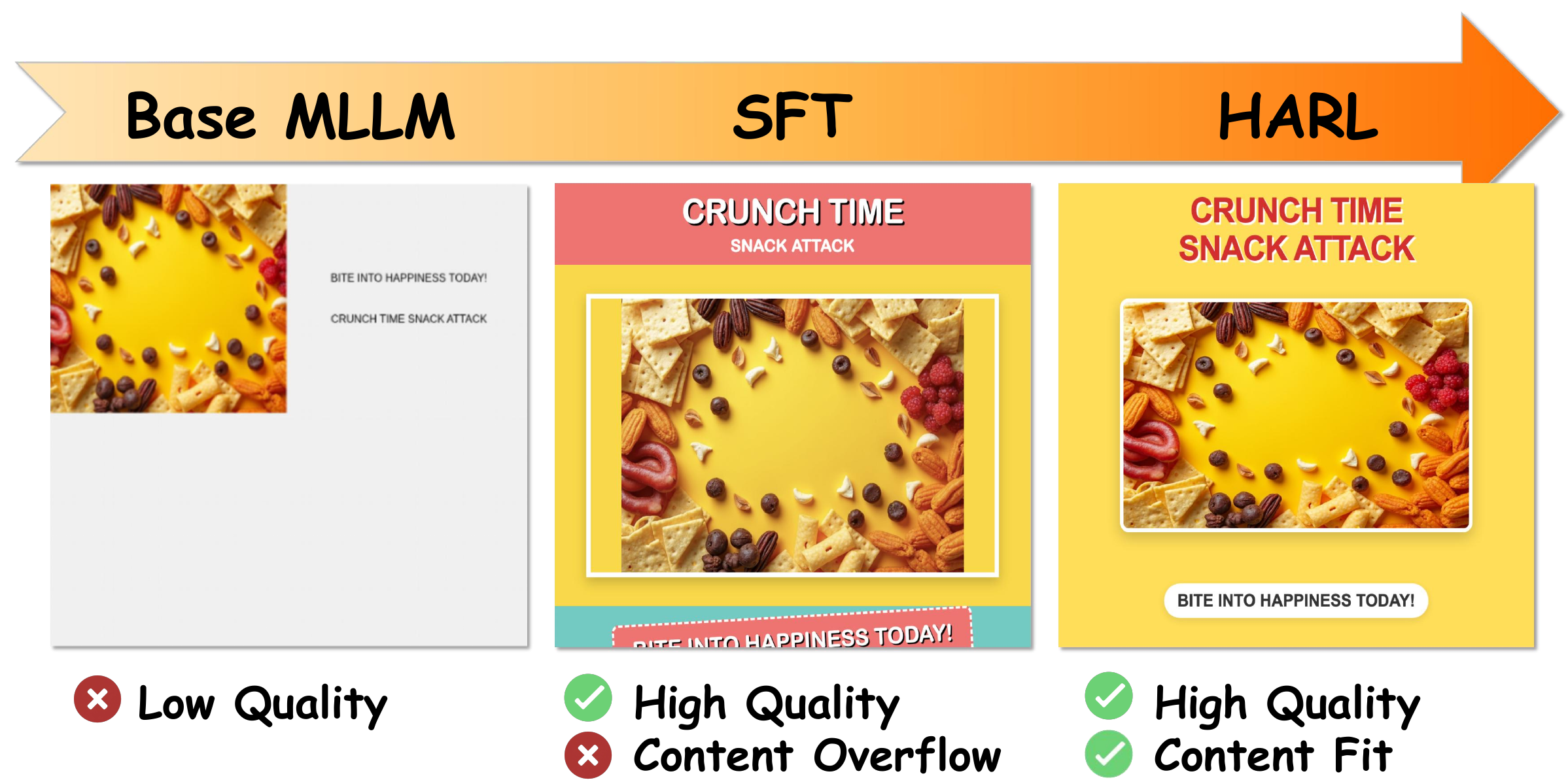}
    \caption{The general multi-modal large language models (MLLMs) produce low-quality documents. Through supervised fine-tuning (SFT) on \dataset{}, the model shows powerful document generation capabilities but still suffers from content overflow. Height-Aware Reinforcement Learning (HARL) is then adopted to overcome this issue while maintaining high visual quality.}
    \label{fig:anydoc_method}
\end{figure}

\subsection{Model Stage I: Supervised Fine-Tuning}
\label{sec:method_sft}

\dataset{} supports various document generation tasks under different input conditions.
In \approach{}, we focus on three representative tasks: intention-to-document (I2D), document derendering (DD), and element-to-document (E2D).
Their generation conditions are textual intentions, screenshots, and elements, respectively (Figure~\ref{fig:teaser}a).
Besides, all tasks take the document dimension (i.e., width and height) as an additional input.
Notably, the application scope of \dataset{} is not limited to the above tasks.
It can also be leveraged for other potential tasks, such as document variation, image-based document generation, document question answering, etc.

We fine-tune a multi-modal large language model to endow it with strong document generation capabilities.
The input-output pairs for each task can be conveniently constructed based on \dataset{}.
For instance, in the E2D setting, the input comprises the document’s width, height, and all elements extracted from the target document, while the corresponding HTML/CSS code serves as the output.
The elements are randomly shuffled to prevent information leakage.
An example of training data is shown in the Appendix.

\subsection{Model Stage II: Height-Aware Reinforcement Learning}
\label{sec:method_rl}

Through supervised fine-tuning (SFT), the model achieves enhanced performance in generating high-quality documents.
However, as shown in Figure~\ref{fig:anydoc_method}, a critical issue is observed in the SFT output: some documents have overflow elements that extend beyond the document boundaries.
As a result, when rendered with the specified height, they appear truncated and exhibit undesirable results.

To this end, we propose Height-Aware Reinforcement Learning (HARL) based on Group Relative Policy Optimization~\cite{shao2024deepseekmath} (GRPO) for post-training.
The motivation stems from GRPO’s group relative advantage mechanism.
For each input query, GRPO samples a group of candidate outputs, producing multiple code implementations of the target document.
Within the group, some implementations may have closer document heights to the target value, i.e., showing less content overflow.
Accordingly, higher rewards should be assigned to such outputs to encourage their generation, while penalizing those with severe overflow.
Built upon this idea, HARL involves the following steps: employing Playwright to render the generated HTML/CSS source code into a document image, obtaining its height $\hat{h}$, and computing the reward $r$ according to the relative value of $\hat{h}$ with respect to the specified height $h$.
Formally, we have
\begin{equation}
\resizebox{0.45\textwidth}{!}{$
r = \max \Biggl( 0,\,
\begin{cases}
1, & \text{if } 1 - \gamma \leq \rho \leq 1, \\[6pt]
1 - (1 - \gamma - \rho) = \gamma + \rho, & \text{if } \rho < 1 - \gamma, \\[6pt]
1 - \alpha \bigl(\rho - 1\bigr), & \text{if } \rho > 1,
\end{cases}
\Biggr).
$}
\end{equation}
Here, $\rho = \frac{\hat{h}}{h}$ represents the height ratio, $\gamma$ and $\alpha$ are the tolerance factor and overflow penalty coefficient.
Notably, we also penalize underflow documents ($\rho < 1 - \gamma$) to prevent reward hacking.
The standard GRPO loss in~\cite{shao2024deepseekmath} is applied for model optimization, further improving model performance in document generation.

\begin{table}[tbp]
\centering
\resizebox{0.48\textwidth}{!}{%
\begin{tabular}{l|ccccccccc}
\toprule
\textbf{Methods} & \textbf{Layout} & \textbf{Image} & \textbf{Typography} & \textbf{Content} & \textbf{Innovation} & \textbf{Ove} & \textbf{Ali} & \textbf{Height} & \textbf{Intention}  \\
\midrule
\rowcolor{gray!15}
\multicolumn{10}{l}{\textit{\textbf{Specialized Models}}} \\
\rowcolor{gray!8}
OpenCOLE & 7.91 & 8.03 & 7.79 & 7.52 & 6.91 & - & - & - & 8.13 \\
\rowcolor{gray!8}
FLUX.1-dev & 7.58 & 7.78 & 6.54 & 5.16 & 6.25 & - & - & - & 6.91 \\
\midrule
\rowcolor{blue!15}
\multicolumn{10}{l}{\textit{\textbf{General MLLMs}}} \\
\rowcolor{blue!8}
InternVL3-78B & 7.62 & 8.16 & 7.26 & 7.37 & 6.31 & 0.0286 & \textbf{0.0014} & 0.323 & 8.02 \\
\rowcolor{blue!8}
GPT-4o & 8.59 & 8.75 & 8.32 & 8.41 & \textbf{7.27} & 0.2540 & 0.0020 & 0.047 & \textbf{8.96} \\
\midrule
\rowcolor{red!15}
\multicolumn{10}{l}{\textit{\textbf{Our Method}}} \\
\rowcolor{red!8}
\approach{} & \textbf{8.64} & \textbf{8.92} & \textbf{8.36} & \textbf{8.44} & 7.21 & \textbf{0.0238} & 0.0053 & \textbf{0.005} & 8.95 \\
\bottomrule
\end{tabular}}
\caption{Quantitative results on the intention-to-document task.}
\label{tab:I2D}
\end{table}


\begin{table}[tbp]
\centering

\resizebox{0.48\textwidth}{!}{%
\begin{tabular}{l|cccccccc}
\toprule
\textbf{Methods} & \textbf{Block} & \textbf{Text} & \textbf{Position} & \textbf{Color} & \textbf{CLIP} & \textbf{Ove} & \textbf{Ali} & \textbf{Height} \\
\midrule
\rowcolor{gray!15}
\multicolumn{9}{l}{\textit{\textbf{Specialized Models}}} \\
\rowcolor{gray!8}
WebSight & 0.838 & 0.902 & 0.755 & 0.797 & 0.796 & 0.0873 & 0.0019 & 0.438 \\
\rowcolor{gray!8}
Design2Code & 0.909 & \textbf{0.989} & 0.732 & 0.651 & 0.758 & 0.3756 & 0.0010 & 0.652 \\
\midrule
\rowcolor{blue!15}
\multicolumn{9}{l}{\textit{\textbf{General MLLMs}}} \\
\rowcolor{blue!8}
InternVL3-78B & 0.910 & 0.952 & 0.788 & 0.757 & 0.792 & \textbf{0.0298} & 0.0012 & 0.134 \\
\rowcolor{blue!8}
GPT-4o & 0.953 & 0.965 & 0.871 & 0.899 & 0.830 & 1.0009 & \textbf{0.0008} & \textbf{0.034} \\
\midrule
\rowcolor{red!15}
\multicolumn{9}{l}{\textit{\textbf{Our Method}}} \\
\rowcolor{red!8}
\approach{} & \textbf{0.965} & 0.986 & \textbf{0.910} & \textbf{0.958} & \textbf{0.835} & 0.0625 & 0.0017 & 0.309 \\
\bottomrule
\end{tabular}}
\caption{Quantitative results on the document derendering task.}
\label{tab:DD}
\end{table}


\section{Experiments}
\label{sec:exp}

\subsection{Setups}
\label{subsec:exp_setup}

\noindent \textbf{Dataset.}
We conduct experiments on the curated \dataset{}, which is split into training, validation, and test sets in an 8:1:1 ratio.
The full training set is used for SFT.
For HARL post-training, the training data is constructed using the most challenging cases.
Specifically, we perform inference on the training set using the SFT model, compute the height deviations from the ground truth values, sort the results, and select the top 20,000 samples as HARL's training set.
Due to the high computational cost of evaluating the full test set, we randomly sample 1,000 instances from it for metric computation in all experiments.

\begin{figure*}[t]
    \centering
    \includegraphics[width=0.99\linewidth]{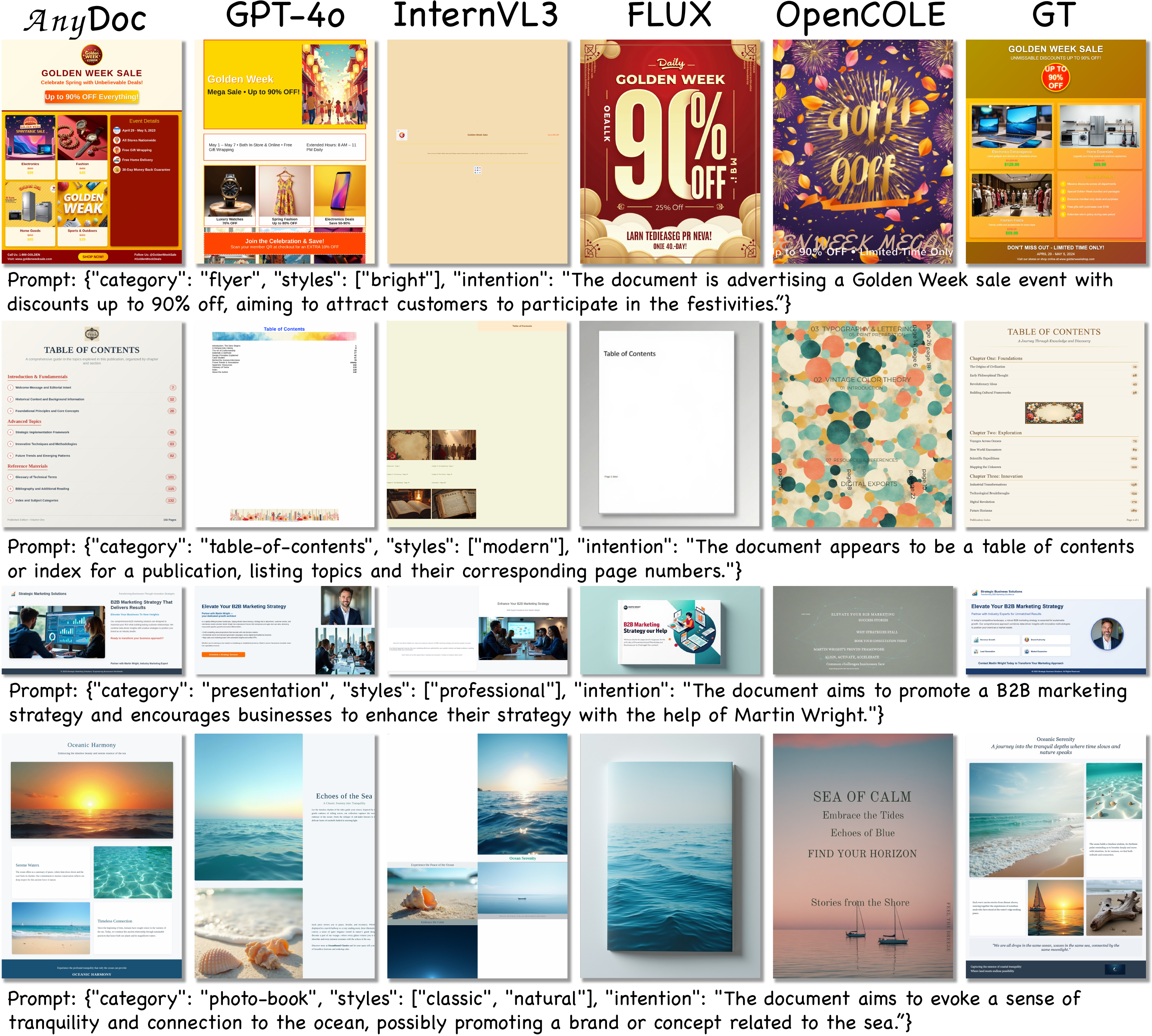}
    \caption{Qualitative results on the intention-to-document task. The corresponding input conditions (i.e., intention, category, and style) are displayed below each example. For reference, the ground truth (GT) documents are also visualized.}
    \label{fig:qualitative_i2d}
\end{figure*}

\vspace{2mm}
\noindent \textbf{Implementation Details.}
We use Qwen2.5-VL-7B-Instruct~\cite{Qwen2.5-VL} as the base model. 
Both SFT and HARL are conducted on a single node with 8 NVIDIA A100 GPUs. 
For SFT, we employ the LLaMA-Factory framework~\cite{zheng2024llamafactory} and apply LoRA fine-tuning~\cite{hu2021loralowrankadaptationlarge} with a rank of 32 to reduce computational overhead given the large training set size. 
The global batch size and learning rate are set to 128 and 1e-4.
For HARL post-training, we adopt the VeRL library~\cite{sheng2024hybridflow}. 
Unlike SFT, we perform full-parameter tuning for more effective policy updates, with a learning rate of 1e-6 and a batch size of 64.
The GRPO rollout number is set to 5.
The KL divergence coefficient and clipping ratio in the GRPO loss function are 0.01 and 0.2, respectively.
The training processes for different tasks are independent.

\begin{table}[tbp]
\centering
\resizebox{0.48\textwidth}{!}{%
\begin{tabular}{l|cccccccc}
\toprule
\textbf{Methods} & \textbf{Layout} & \textbf{Image} & \textbf{Typography} & \textbf{Content} & \textbf{Innovation} & \textbf{Ove} & \textbf{Ali} & \textbf{Height} \\
\midrule
\rowcolor{gray!15}
\multicolumn{9}{l}{\textit{\textbf{Specialized Models}}} \\
\rowcolor{gray!8}
LaDeCo & 7.58 & 8.25 & 7.11 & 7.32 & 6.71 & 0.7150 & 0.0022 & \textbf{0} \\
\midrule
\rowcolor{blue!15}
\multicolumn{9}{l}{\textit{\textbf{General MLLMs}}} \\
\rowcolor{blue!8}
InternVL3-78B & 7.44 & 8.05 & 7.04 & 7.16 & 6.42 & \textbf{0.0124} & \textbf{0.0008} & 0.934 \\
\rowcolor{blue!8}
GPT-4o & 8.09 & 8.38 & 7.90 & 7.79 & 6.87 & 0.2520 & 0.0017 & 0.212 \\
\midrule
\rowcolor{red!15}
\multicolumn{9}{l}{\textit{\textbf{Our Method}}} \\
\rowcolor{red!8}
\approach{} & \textbf{8.62} & \textbf{8.73} & \textbf{8.39} & \textbf{8.34} & \textbf{7.36} & 0.0494 & 0.0021 & 0.085 \\
\bottomrule
\end{tabular}}
\caption{Quantitative results on the element-to-document task.}
\label{tab:E2D}
\end{table}

\vspace{2mm}
\noindent \textbf{Compared Baselines.}
For all three tasks, we establish common baselines using general multi-modal large language models (MLLMs) in the zero-shot setting, including GPT-4o~\cite{achiam2023gpt} and InternVL3-78B~\cite{zhu2025internvl3}.
In addition, each task also has specialized baselines.
(1) I2D: \approach{} is compared with OpenCOLE~\cite{inoue2024opencole} and the state-of-the-art text-to-image model FLUX.1-dev~\cite{flux2024}.
(2) DD: We choose two recent screenshot-to-code approaches, WebSight~\cite{laurençon2024unlockingconversionwebscreenshots} and Design2Code~\cite{si2025design2codebenchmarkingmultimodalcode}, as the baseline models.
(3) E2D: We compare against the layered element-to-design framework LaDeCo~\cite{lin2025elements}.
More baseline details can be found in the supplementary materials.

\vspace{2mm}
\noindent \textbf{Evaluation Metrics.}
We follow the baseline models to set up the evaluation metrics.
(1) I2D: Following OpenCOLE~\cite{inoue2024opencole}, we employ MLLM-based judge scores to assess the generation quality across five dimensions: Layout, Image, Typography, Content, and Innovation.
The scoring MLLM is InternVL3-78B~\cite{zhu2025internvl3}.
In addition, we introduce an Intention metric to measure the semantic consistency between the generated document and the input intention, which is also evaluated by the same MLLM.
(2) DD: Following Design2Code~\cite{si2025design2codebenchmarkingmultimodalcode}, we adopt Block, Text, Position, Color, and CLIP metrics to evaluate the derendering quality and accuracy.
(3) E2D: Following LaDeCo~\cite{lin2025elements}, we use the same five MLLM-based metrics as in the I2D task to score the generated documents.
For all tasks, we further include Overlap (Ove) and Alignment (Ali) metrics to evaluate layout quality, and a Height metric to quantify the degree of content overflow.
For Overlap, Alignment, and Height metrics, lower is better. For other metrics, higher is better.

\subsection{Main Results}

\noindent \textbf{Quantitative Results.}
The quantitative results for the I2D, DD, and E2D tasks are shown in Tables~\ref{tab:I2D},~\ref{tab:DD}, and~\ref{tab:E2D}.
For I2D, \approach{} surpasses or matches the baselines on nearly all metrics, demonstrating its ability to generate high-quality and semantically aligned documents from textual intentions.
For DD, \approach{} achieves strong results across the Block, Text, Position, Color, and CLIP metrics, indicating high derendering accuracy. 
Although the Height metric is not ideal, i.e., occasionally showing overflow, this is expected given that our model size is relatively small compared to InternVL3 and GPT-4o.
In fact, specialized derendering models such as WebSight and Design2Code perform even worse on Height. 
We believe that this issue can be alleviated by scaling up the model.
For E2D, \approach{} significantly outperforms GPT-4o across the five MLLM-based metrics and Height, confirming its ability to generate high-quality documents from a set of input elements.

\begin{figure*}[t]
    \centering
    \includegraphics[width=0.99\linewidth]{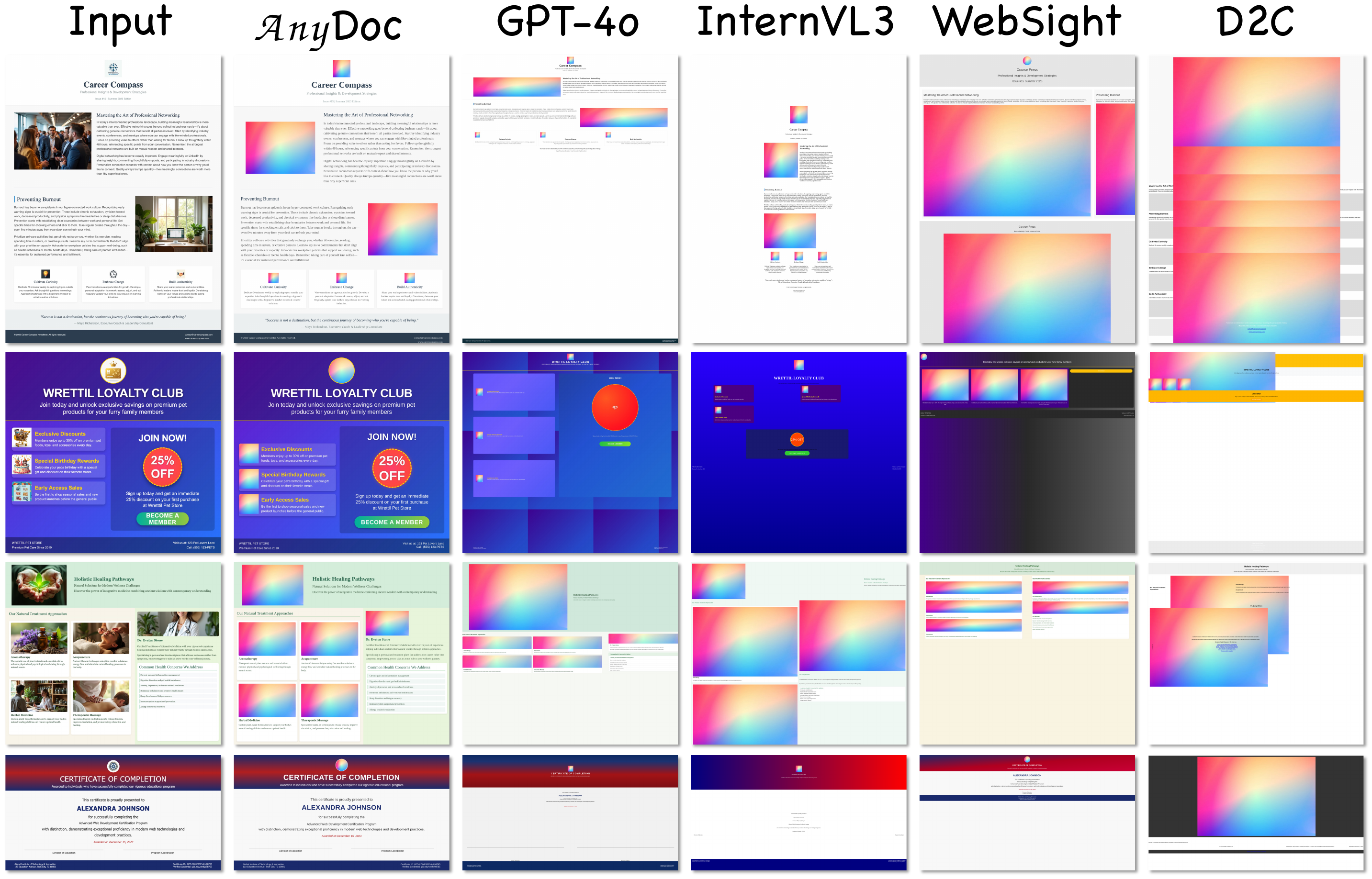}
    \caption{Qualitative results on the document derendering task. D2C is Design2Code. Gradient images are used for image placeholders. \approach{} produces highly accurate derendering results that closely resemble the input documents.}
    \label{fig:qualitative_dd}
\end{figure*}
\begin{table*}[tbp]
\centering

\resizebox{0.93\textwidth}{!}{%
\begin{tabular}{ccc|cccccccc}
\toprule
\textbf{Format} & \textbf{Data} & \textbf{HARL} & \textbf{Layout} & \textbf{Image} & \textbf{Typography} & \textbf{Content} & \textbf{Innovation} & \textbf{Ove} & \textbf{Ali} & \textbf{Height} \\
\midrule
HTML/CSS & 10K & \ding{55} & 8.47 & 8.56 & 8.29 & 8.22 & 7.21 & 0.0542 & 0.0015 & 0.299 \\
HTML/CSS & 50K & \ding{55} & 8.48 & 8.67 & 8.28 & 8.20 & 7.25 & 0.0822 & 0.0018 & 0.493 \\
\midrule
Flat & Full & \ding{55} & 8.43 & 8.55 & 8.08 & 8.15 & 7.21 & 0.2627 & 0.0015 & 0 \\
HTML/CSS & Full & \ding{55} & 8.50 & 8.61 & 8.30 & 8.27 & 7.25 & 0.1091 & 0.0017 & 0.392 \\
\midrule
HTML/CSS & Full & \ding{51} & 8.62 & 8.73 & 8.39 & 8.34 & 7.36 & 0.0494 & 0.0021 & 0.085 \\
\bottomrule
\end{tabular}}
\caption{Ablation studies results. We conduct these experiments on the E2D task. Flat denotes the flat coordinate representation in~\cite{lin2025elements}.}
\label{tab:E2D_ablation}
\end{table*}



\begin{figure*}[t]
    \centering
    \includegraphics[width=0.99\linewidth]{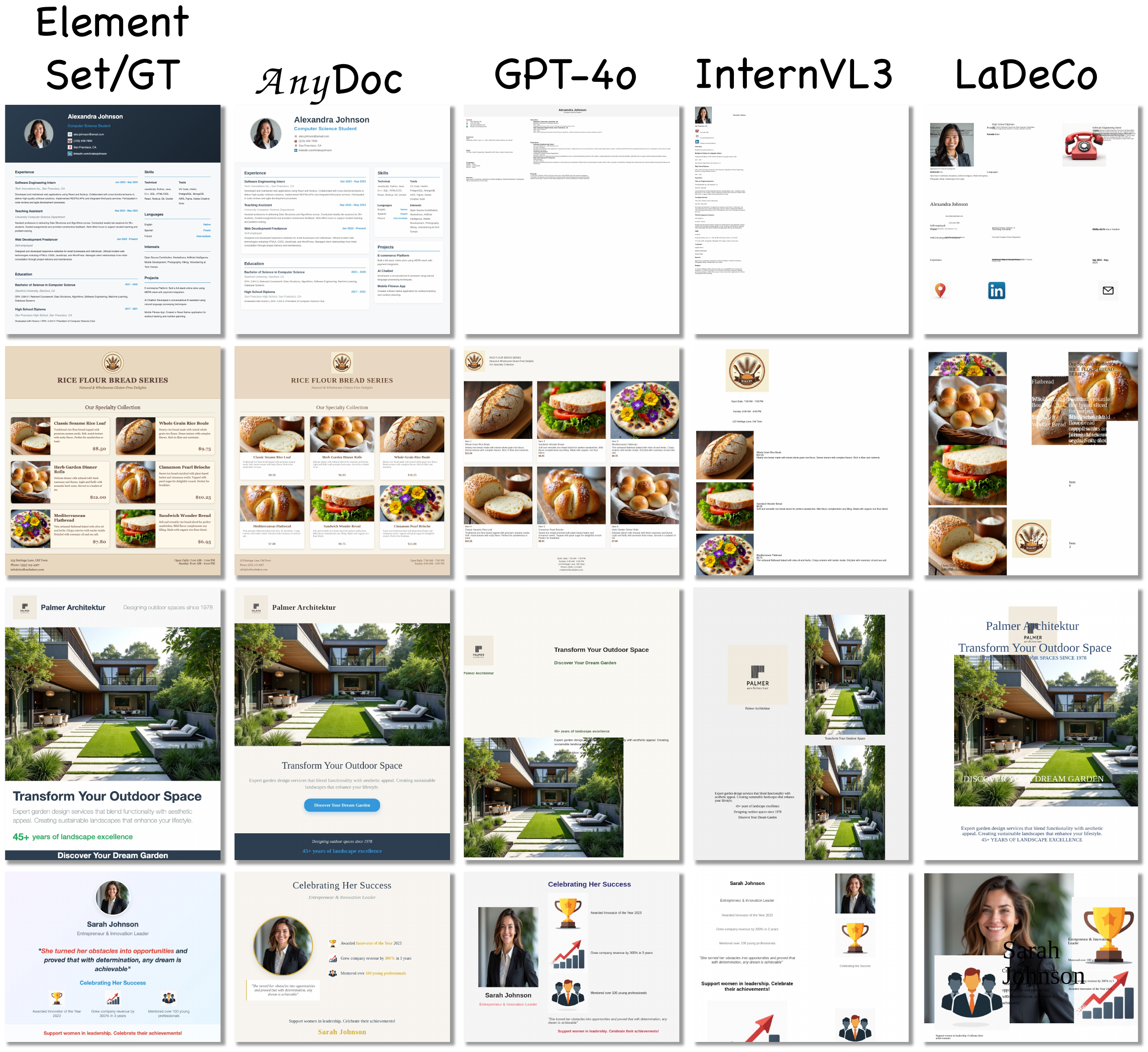}
    \caption{Qualitative results on the element-to-document task. The elements in the ground truth documents serve as the model input. \approach{} combines the input individual elements into a high-quality document output.}
    \label{fig:qualitative_e2d}
\end{figure*}

\vspace{2mm}
\noindent \textbf{Qualitative Results.}
We visualize the qualitative comparison results of the three tasks in Figures~\ref{fig:qualitative_i2d},~\ref{fig:qualitative_dd}, and~\ref{fig:qualitative_e2d}.
\approach{} demonstrates remarkable document generation capabilities, producing high-quality, visually appealing outputs that effectively achieve the tasks.
Specifically, in Figure~\ref{fig:qualitative_i2d}, \approach{} accurately captures the input semantics and creates documents that align well with the specified categories, styles, and intentions.
In Figure~\ref{fig:qualitative_dd}, it successfully reconstructs the original documents with superior fidelity in layout structure, color schemes, and text recognition.
In the E2D task of Figure~\ref{fig:qualitative_e2d}, \approach{} effectively combines the provided elements into coherent documents while maintaining proper visual hierarchy and design principles.

In contrast, baseline methods exhibit notable limitations such as poor element arrangement, severe overlap issues, unnatural spacing, etc.
They also fail to effectively accomplish these three tasks.
For instance, in Figure~\ref{fig:qualitative_dd}, WebSight and Design2Code struggle with layout fidelity during reconstruction.
LaDeCo encounters substantial difficulties when handling structurally complex documents in Figure~\ref{fig:qualitative_e2d}.
Besides, both FLUX and OpenCOLE in Figure~\ref{fig:qualitative_i2d} produce raster image outputs, which restricts their applicability in editable scenarios.

\subsection{Ablation Studies}

We conduct a series of ablation studies to examine the contribution of the key components in \approach{}.
First, we remove HARL from \approach{}. 
As shown in Table~\ref{tab:E2D_ablation}, this leads to a significant degradation in the Height metric, indicating that our post-training procedure effectively reduces content overflow.
The five MLLM-based scores also decline, further suggesting that the overall document quality is impaired without HARL.
Second, we investigate the effect of proposed HTML/CSS document representation. 
We implement a variant following existing work~\cite{lin2025elements} that uses a flat coordinate representation.
The results in Table~\ref{tab:E2D_ablation} show a substantial deterioration in the Overlap metric, i.e., clear layout quality issues. 
This validates that coordinate-based representations are suboptimal for document representation, as they require extensive geometric computation and provide limited structural expressiveness.
Third, we analyze the effect of the synthesized dataset \dataset{}. 
When increasing the training data size from 10K to 50K and then to the full dataset, the quantitative metrics remain stable, demonstrating the robustness of \approach{}. 
Moreover, even with only 10K samples, the model already outperforms GPT-4o across most metrics in Table~\ref{tab:E2D}, highlighting the effectiveness of our curated dataset for document generation.

\section{Conclusion}

This work presents \approach{} to advance document generation.
It handles multiple document generation tasks across a diverse range of document categories.
To achieve this, \approach{} first introduces HTML/CSS as a unified document representation.
Then, it includes a supervised fine-tuning (SFT) process to acquire basic document capabilities, and a height-aware reinforcement learning (HARL) post-training process to mitigate content overflow.
Besides, we develop an automated data synthesis pipeline to curate \dataset{}, which contains 265,206 high-quality documents that cover 111 categories and 32 styles.
Quantitative and qualitative results demonstrate the effectiveness of \approach{}.
In the future, we will extend it to support more document generation tasks, and explore strategies to more thoroughly address content overflow, further improving the performance and usability of \approach{}.

\newpage
{
    \small
    \bibliographystyle{ieeenat_fullname}
    \bibliography{main}
}

\clearpage
\onecolumn
\setcounter{figure}{7}
\setcounter{table}{4}
\appendix

\section{Prompts Used in the Dataset Construction Pipeline}
\label{sec:more_prompts}

\subsection{Intention and Description Annotation}

\begin{mdframed}[backgroundcolor=yellow!10, linecolor=gray]
\setlength{\parindent}{0pt}
\footnotesize
You are an excellent document analyst.
I will give you a document image.
Your task is to carefully (1) infer its intention, and (2) provide a detailed description of its content.

Instructions:

- For the intention, provide 1-2 high-level sentences about the document's purpose, goal, or the message it aims to convey. Please don't include visual details.

- For the description, write 2-3 sentences giving a clear, objective, and factual content description of the document. Focus on what is present in the document, without inferring beyond the visible elements.

- Please strictly follow this JSON format for your output, {``intention'': `` '', ``description'': `` ''}.

- Do not include \texttt{```} or \texttt{```}json in your answer.

- Do not explain or justify your answer.
\end{mdframed}

\subsection{HTML/CSS Document Code Generation}
\label{sec:code_generation_prompt}

\begin{mdframed}[backgroundcolor=orange!10, linecolor=gray]
\setlength{\parindent}{0pt}
\footnotesize
You are a professional graphic designer.
Your task is to create a high-quality document using HTML and CSS.

Instructions:

- Create the specified document as HTML/CSS.

- Do not use Javascript.

- Do not use any frameworks or libraries.

- Besides necessary structural tags (\texttt{<html>}, \texttt{<head>}, \texttt{<body>}, etc.), use only \texttt{<div>}, \texttt{<img>}, and \texttt{<p>} tags.

- All text must appear in a \texttt{<p>} tag.

- All images/icons/logos must appear in an \texttt{<img>} tag. 

- The \texttt{<img>} tag must have both src and alt fields.

- For the src of all \texttt{<img>} tags, use only this source: ``https://picsum.photos/W/H'', by replacing ``W'' and ``H'' in the URL with the desired width and height. For example, ``https://picsum.photos/200/300'' for an image of width=200 and height=300. Make sure the width/height attributes of the \texttt{<img>} tag match the dimensions in the URL.

- In the alt text of all \texttt{<img>} tags, write a detailed description of what the image should be. This description will be used to search or generate an appropriate image, so the more detailed, the better!

- Do not use placeholder shapes for icons or logos.

- Do not include an image in ``background: url(...)''.

- Make the \texttt{<body>} tag a fixed size based on the document size.

- \texttt{<body>} and \texttt{<div>} tags must use either ``display: flex'' or ``display: grid''.

- Grids may only be used if there are at least 2 columns and 2 rows.

- Flex containers should always set ``flex-wrap: nowrap''. Flex direction should be ``row'' or ``column''. Do not use reverse. 

- Do not use flex item reordering.

- Prefer to use ``row-gap'', ``column-gap'', ``gap'', ``justify-content'', and ``align-items'' to control spacing and alignment of elements.

- Prefer to use ``flex-basis'', ``flex-grow'', and ``flex-shrink'' to control the size of flex items.

- All elements should have ``box-sizing: border-box''. Do not use padding unless an element has a visible border, i.e. prefer margin to padding.

- Prefer specifying dimensions in percentages rather than pixels unless the dimension is small.

- No absolute positioning of elements.

Tips!!! Use these principles to create an aesthetically pleasing document:

* Proper content alignment is key.

* Use hierarchy to help focus your design.

* Leverage contrast to accentuate important design elements.

* Choose fonts and font sizes that are stylistically appropriate.

* Not all text needs to be the same font or size.

* Use repetition to create a cohesive look.

* Consider proximity when organizing your graphic elements.

* Make sure your design is balanced.

* Carefully select colors that are diverse, visually appealing, and appropriate for the theme.

* Leave negative space.

\end{mdframed}

\newpage

\section{Supervised Fine-Tuning Input-Output Examples}
\label{sec:data_example}

\begin{mdframed}[backgroundcolor=blue!10, linecolor=gray]
\setlength{\parindent}{0pt}
\footnotesize
\textbf{I2D Input}

{``width'': 3300, ``height'': 2071, ``category'': ``planner'', ``styles'': [``natural''], ``intention'': ``The document is intended to provide a structured template for outlining the details of a course, including the subject, days, teacher, and additional course details.''}

\textbf{DD Input}

{``width'': 3300, ``height'': 2071, ``screenshot'': ``\texttt{<image>}''}

\textbf{E2D Input}

{``width'': 3300, ``height'': 2071, ``texts'': [``Dr. Evelyn Green'', ``Wednesday'', ``Materials'', ``Additional Information'', ``Duration'', ``Prerequisites'', ``Office Hours: Tue/Thu 1-3 PM'', ``12 Weeks'', ``Subject'', ``evelyn.green@university.edu'', ``2:00 PM - 4:00 PM'', ``Course Details'', ``10:00 AM - 12:00 PM'', ``Botany \& Ecology'', ``Advanced'', ``Advanced Botany Studies'', ``Learning Outcomes'', ``Intro to Biology'', ``Course Planner Template - Natural Theme'', ``Schedule'', ``Upon completion of this course, students will be able to analyze plant physiological processes, evaluate ecosystem dynamics, conduct botanical research using scientific methods, and apply knowledge of plant biology to environmental conservation efforts.'', ``10:00 AM - 12:00 PM'', ``Greenhouse Lab B'', ``Monday'', ``Professor of Botany'', ``Lab Manual \& Field Kit'', ``Exploring Plant Life and Ecosystems'', ``Level'', ``This comprehensive course explores the intricate relationships between plant life and their ecosystems. Students will examine plant physiology, taxonomy, and ecological roles through hands-on laboratory work and field studies. Topics include photosynthesis mechanisms, plant adaptations, biodiversity conservation, and sustainable gardening practices. The course emphasizes critical thinking and scientific methodology through research projects and collaborative learning experiences.'', ``Projects \& Exams'', ``Fall Semester 2023'', ``4.0'', ``Course Description'', ``Instructor'', ``Assessment'', ``Credits'', ``Location'', ``Office: Science Building, Room 304'', ``Friday''], ``images'': {``image\_0'': ``\texttt{<image>}'', ``image\_1'': ``\texttt{<image>}'', ``image\_2'': ``\texttt{<image>}'', ``image\_3'': ``\texttt{<image>}'', ``image\_4'': ``\texttt{<image>}'', ``image\_5'': ``\texttt{<image>}'', ``image\_6'': ``\texttt{<image>}'', ``image\_7'': ``\texttt{<image>}''}}

\textbf{Common Output}

\begin{lstlisting}
<!DOCTYPE html>
<html>
  <head>
    <style>
        * {
            box-sizing: border-box;
            margin: 0;
        }

        body {
            width: 3300px;
            height: 2071px;
            display: flex;
            flex-direction: column;
            ...
        }

        .header {
            height: 20%;
            display: flex;
            flex-direction: row;
            align-items: center;
            ...
        }
        ...
    </style>
  </head>
  <body>
    <div class="header">
      <div class="logo-container">
        <!-- I2D or DD img tag -->
        <img src="https://picsum.photos/200/200" alt="Natural green leaf logo representing growth and learning">  
        <!-- E2D img tag -->
        <img src="image_6_200x200.png" alt="Natural green leaf logo representing growth and learning">  
      </div>
      ...
    </div>
    ...
  </body>
</html>

\end{lstlisting}
\end{mdframed}

\section{More Baseline Implementation Details}
\label{sec:more_baseline_details}

\paragraph{GPT-4o and InternVL3-78B.}
We use these two general-purpose MLLMs in the zero-shot setting.
The prompt is almost the same as the code generation prompt used in data construction (Section~\ref{sec:code_generation_prompt}), with the addition of task-specific instructions tailored to each generation task. 
The instructions are:
(1) I2D: \textit{Generate a high-quality document that accurately reflects the specified category, style, and intention.}
(2) DD: \textit{Convert the input document screenshot into corresponding HTML/CSS code that faithfully reproduces the visual layout and content.}
(3) E2D: \textit{Synthesize a coherent document by incorporating all provided elements exactly once, ensuring proper layout and visual harmony.}

\paragraph{FLUX.1-dev.}
We also use FLUX.1-dev as one of the baselines for I2D.
The text prompt is \textit{A flat, digital, single page {category} document in a top-down view, clearly presenting text and image content according to the intention: {intention}. Use {styles} styles.}

\paragraph{LaDeCo.}
LaDeCo serves as the baseline for E2D, employing a five-layer hierarchical framework consisting of background, underlay, image, text, and embellishment layers. 
However, as illustrated in Section~\ref{sec:data_example}, our task formulation focuses on two element types: image and text.
Hence, we adapt LaDeCo by utilizing only its image and text layers while leaving the remaining three layers empty (containing no elements) for fair comparison.

\newpage

\section{Dataset Statistics}

\begin{figure}[h]
    \centering
    \includegraphics[width=\linewidth]{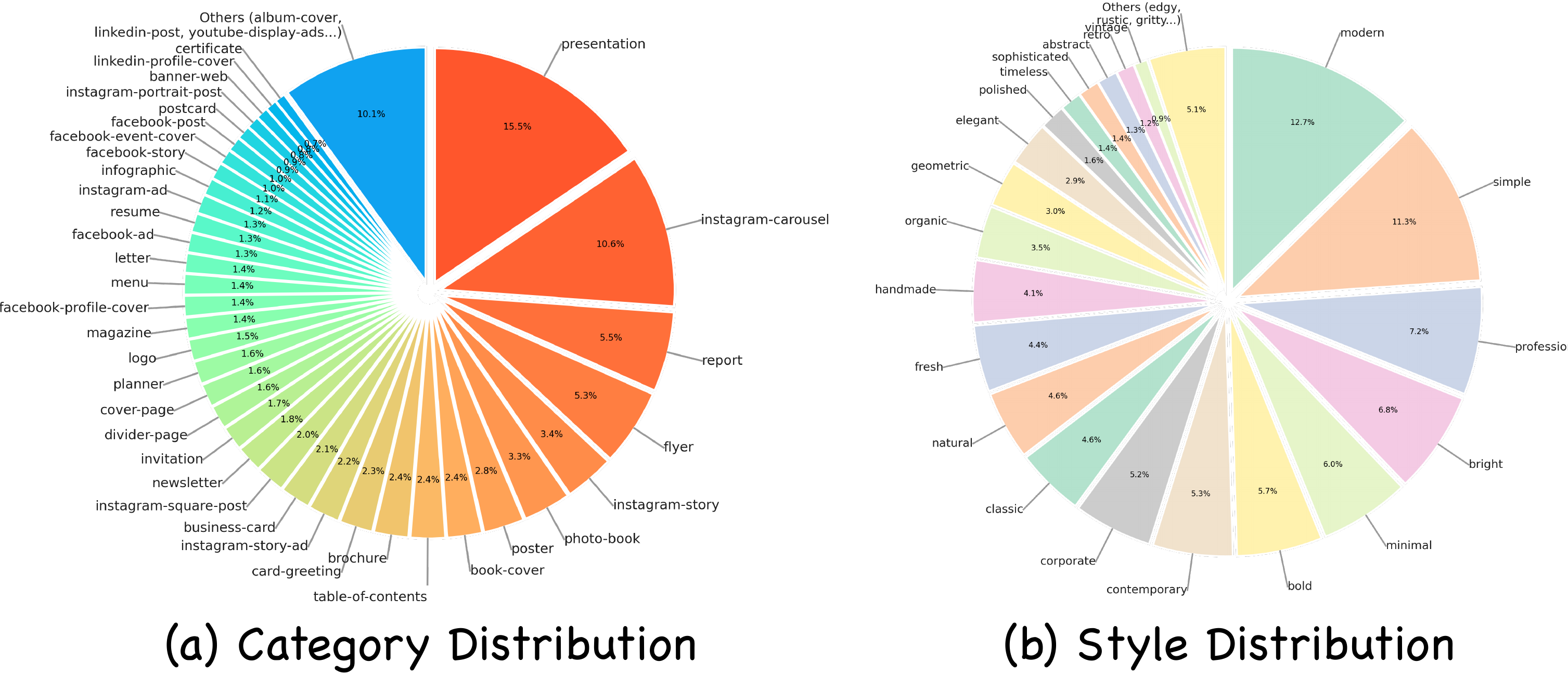}
    \caption{Distributions of document categories and styles in \dataset{}.}
    \label{fig:dataset_pie}
\end{figure}

\begin{figure}[h]
    \centering
    \includegraphics[width=\linewidth]{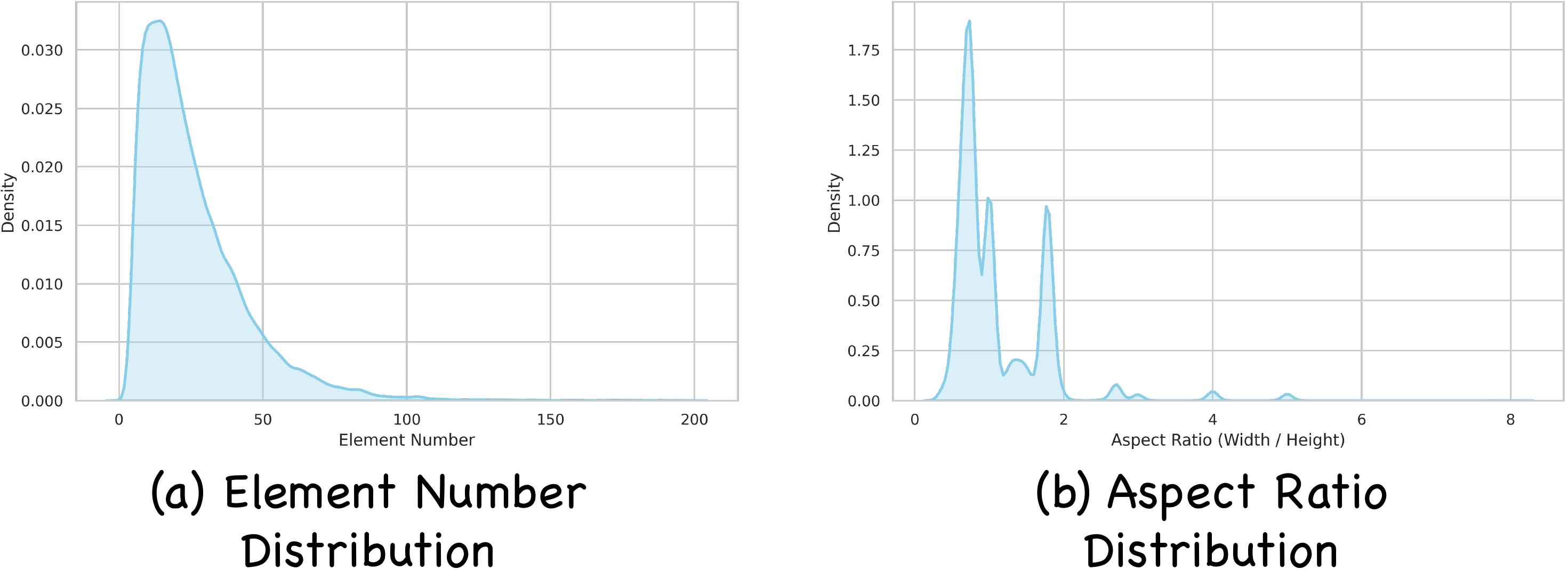}
    \caption{Distributions of document element count and document aspect ratio in \dataset{}.}
    \label{fig:dataset_KDE}
\end{figure}

Figure~\ref{fig:dataset_pie} visualizes the distributions of document categories and styles within \dataset{}.
The element count and document aspect ratio distributions are shown in Figure~\ref{fig:dataset_KDE}.
From the figures, we can see that \dataset{} covers a broad range of document categories, styles, layout complexities, and aspect ratios.

\section{More Qualitative Results}

\begin{figure}
    \centering
    \includegraphics[width=0.62\linewidth]{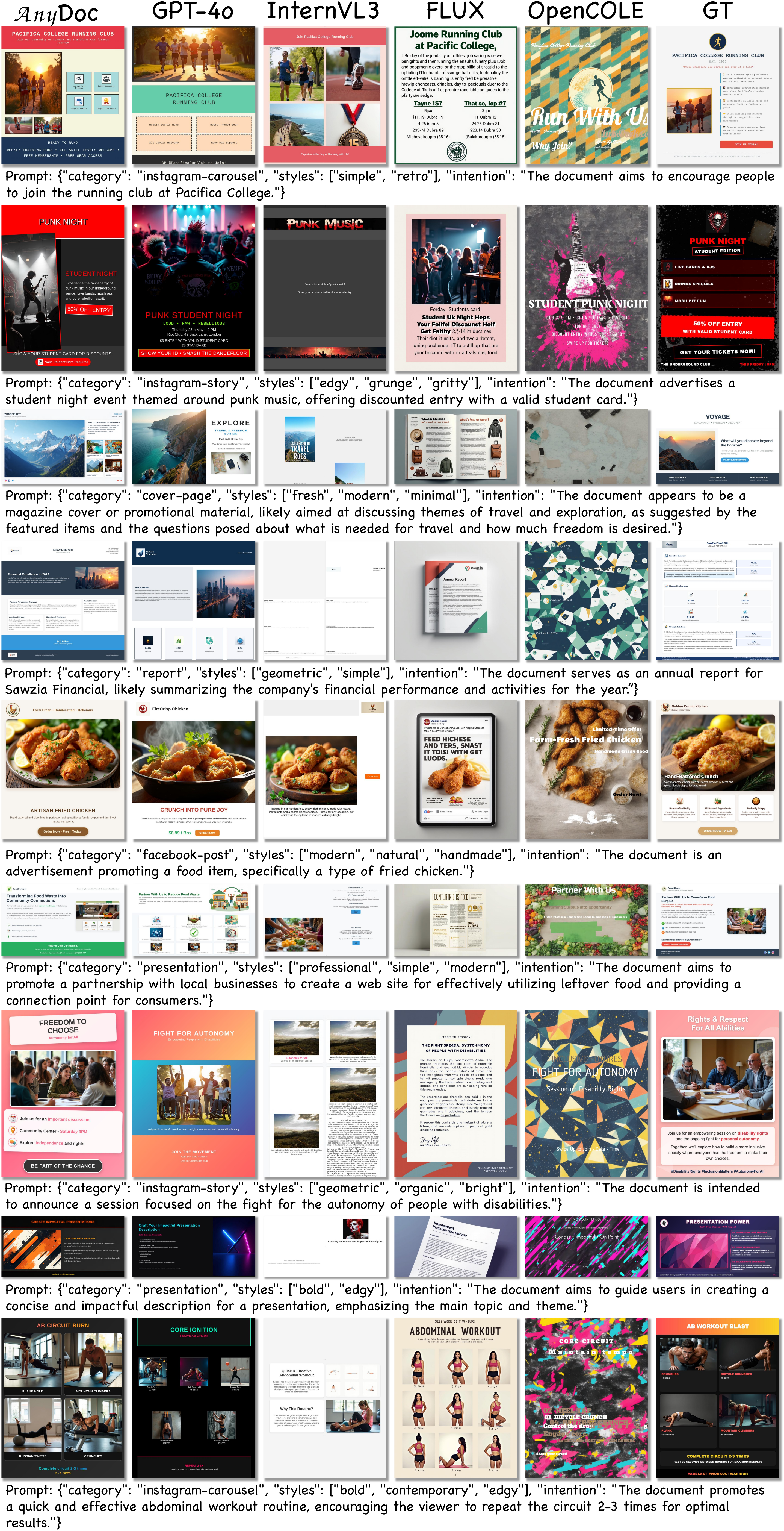}
    \caption{More qualitative results on the intention-to-document task.}
    \label{fig:I2D_supp}
\end{figure}

\begin{figure}
     \centering
     \includegraphics[width=0.75\linewidth]{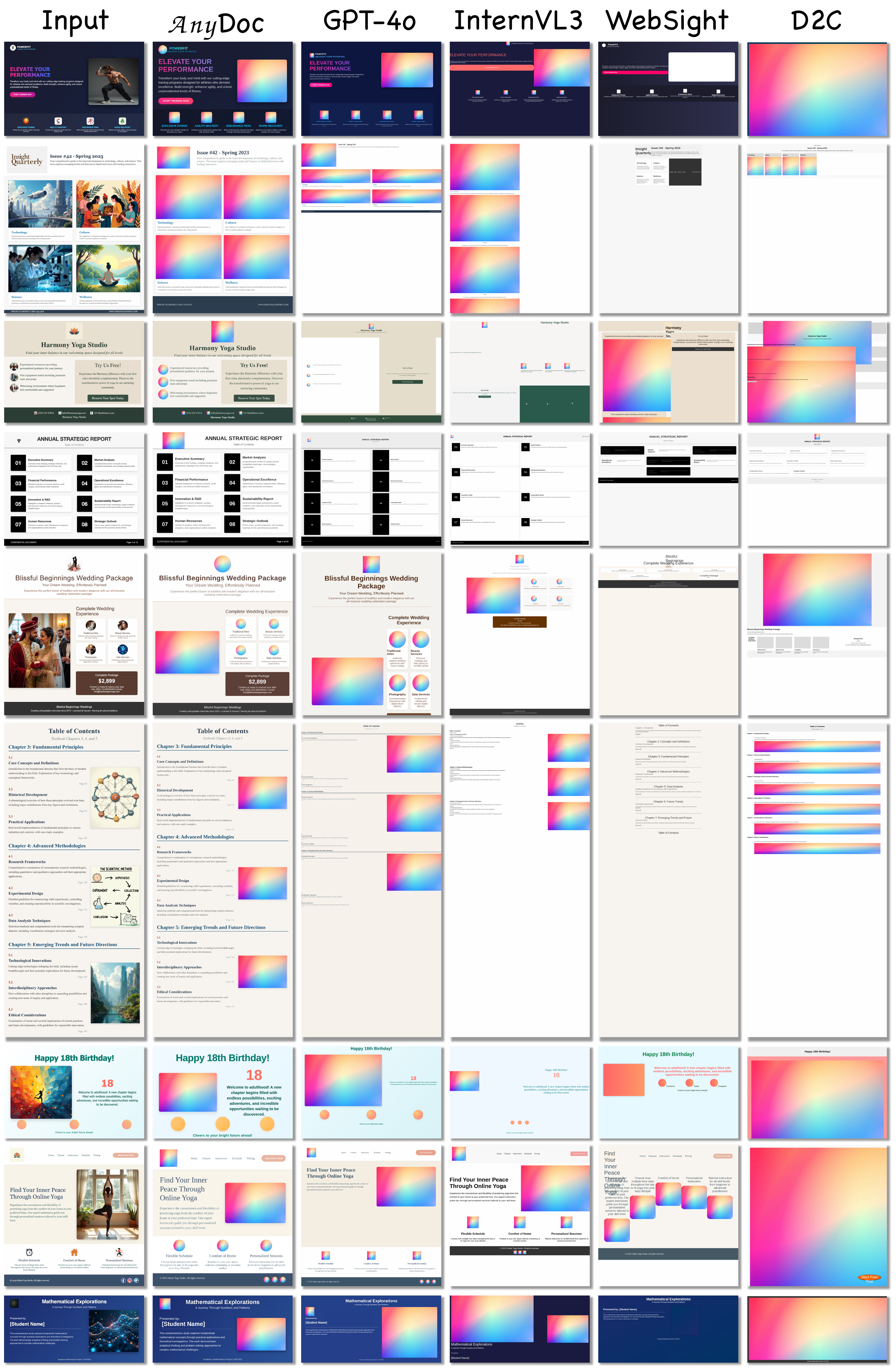}
     \caption{More qualitative results on the document derendering task.}
     \label{fig:DD_supp}
\end{figure}

\begin{figure}
    \centering
    \includegraphics[width=0.62\linewidth]{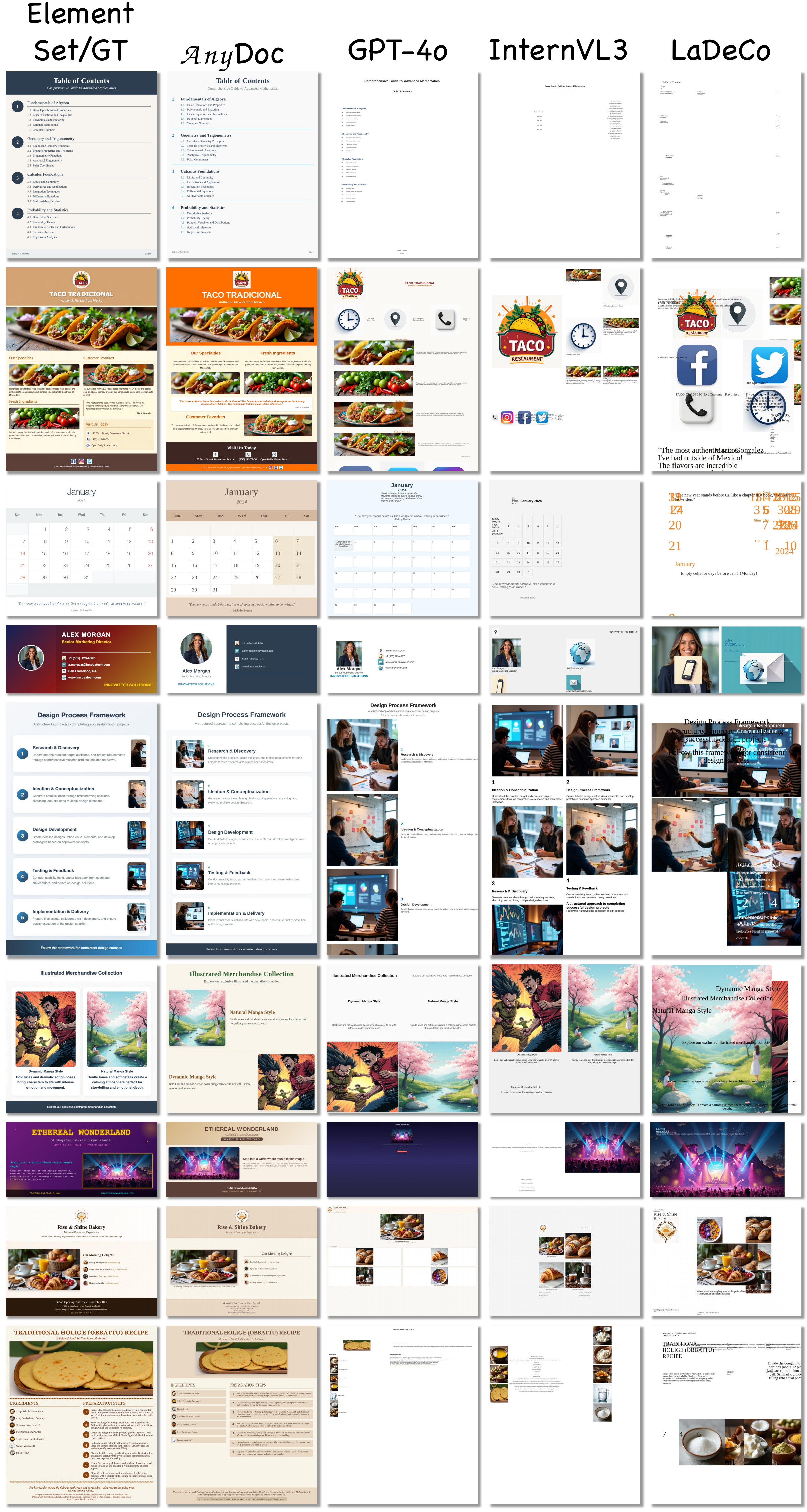}
    \caption{More qualitative results on the element-to-document task.}
    \label{fig:E2D_supp}
\end{figure}

In Figures~\ref{fig:I2D_supp},~\ref{fig:DD_supp},~\ref{fig:E2D_supp}, we show more qualitative comparison of the I2D, DD, and E2D tasks, respectively.
\approach{} exhibits remarkable document generation performance compared to the baseline models.

\end{document}